\documentclass[11pt,a4paper]{article}
\usepackage[hyperref]{acl2020}
\usepackage{times}
\usepackage{latexsym}

\usepackage{microtype}

\usepackage{graphicx}

\usepackage{multirow}
\usepackage{booktabs}

\usepackage[flushleft]{threeparttable}

\usepackage{float}

\aclfinalcopy 

\title{Assessing Post-editing Effort in the English-Hindi Direction}

\author{Arafat Ahsan, Vandan Mujadia \and Dipti Misra Sharma\\
  IIIT, Hyderabad, India \\
  \texttt{\string{arafat.a, vandan.mu\string}@research.iiit.ac.in}\\ \texttt{dipti@iiit.ac.in} \\
  }

\date{}

\begin{document}
\maketitle
\begin{abstract}
We present findings from a first in-depth post-editing effort estimation study in the English-Hindi direction along multiple effort indicators. We conduct a controlled experiment involving professional translators, who complete assigned tasks alternately, in a translation from scratch and a post-edit condition. We find that post-editing reduces translation time (by 63\%), utilizes fewer keystrokes (by 59\%), and decreases the number of pauses (by 63\%) when compared to translating from scratch. We further verify the quality of translations thus produced via a human evaluation task in which we do not detect any discernible quality differences.
\end{abstract}

\section{Introduction}
Translation workflows that are based on post-editing of Machine Translation  output are being increasingly adopted in the industry \citep{gaspari2015survey}. Gains that accrue from a post-editing based workflow, measured over multiple post-editing effort indicators, have been reported to be considerably significant by a number of previous studies over multiple language combinations \citep{plitt2010productivity, c-m-de-sousa-etal-2011-assessing, green2013efficacy}. But to extend post-editing beyond its current silos it is imperative to put new and less-studied language pairs under the lens to make a case for wider adoption via empirically backed evidence.\footnote{\citet{gaspari2015survey}'s survey reveals a heavy skew towards English and other European language combinations.}

Post-editing effort is often quantified across three different dimensions, each focusing in turn on a different aspect of post-editing behaviour \citep{krings2001repairing}. The dimensions studied are the following: \textit{Temporal}-- understood as the time taken to complete a translation task, often reported per segment or word; \textit{Technical}-- estimate of the physical labour of the translation activity, measured in terms of keystrokes logged or edit operations performed; and \textit{Cognitive}-- an indirect estimate of the extent of cognitive processes underlying the translation task, inferred from keylogging pause or eye-tracking data as it is not possible to observe these directly \citep{moorkens2015correlations}.

If it can be shown that post-editing machine translation output is temporally efficient, technically less laborious, and cognitively less demanding, then it can be recommended as the default workflow for large translation jobs. But this first calls for a comparison between machine translation based post-editing  behaviour (henceforth PE) and unaided human translation from scratch (henceforth HT). Thus, the research questions that we pose are the following:

\begin{itemize}
  \item Is post-editing effort as measured on temporal, technical and cognitive dimensions \textit{lesser} in the PE condition than the HT condition for the English-Hindi direction?
  \item Is the quality of post-edited segments \textit{equal to} translated segments as ascertained by human raters?
  \item Do automatic MT evaluation metrics \textit{correlate} with PE effort indicators, when both measured at the segment level?
\end{itemize}

Most of this paper will focus on answering the first question in some detail. We are equally interested in the other two as well, but will only be presenting some initial results from a first attempt at tackling them.

The rest of this paper is organized as follows: Section \ref{related} discusses some past studies including those that have previously studied the English-Hindi PE direction. In Section \ref{setup} we detail our experimental setup. Section \ref{results} presents our results and analysis and in Section \ref{future} we draw our conclusions and sketch the outlines of our future work.

\section{Related Work}
\label{related}

We now take a more detailed look at some of the past efforts towards contrasting the two settings. \citet{plitt2010productivity} compared HT and PE when translating from English into 4 European languages (French, Italian, German, and Spanish) and reported an overall productivity gain of 74\% which  converted into time savings of 43\%. They also observed a 70\% reduction in keyboard time and 31\% in pause time for the PE setting. \citet{c-m-de-sousa-etal-2011-assessing} also report PE to be 40\% faster than HT in the English-Portuguese direction when translating movie subtitles.

Other studies however \citep{laubli2013assessing}, have reported more modest gains, with estimated time savings of 15--20\% when translating between a European language pair (German-French) within a \textit{realistic} translation environment.\footnote{Experimental settings for these studies may deploy  specialized interfaces for accurate measurements or make use of environments already familiar to professional translators.} \citet{garcia2011translating} also finds only marginal productivity gains when studying the English-Chinese pair and additionally reports an impact of directionality when source and target languages are switched.

All of these earlier experiments, however, were based on the output of Phrase based Statistical Machine Translation (PBSMT) systems. With Neural Machine Translation (NMT) and its subsequent iterations being the current state of the art and outperforming PBSMT \citep{bahdanau2014neural, vaswani2017attention, castilho2018evaluating}, this shift in technology paradigm from PBSMT to NMT must then be addressed in post-editing studies as well.

\citet{laubli2019post} conduct such a study, this time utilizing the output of an NMT system to compare  PE with HT in the German-French and German-Italian translation directions.\footnote{The HT condition is aided by a domain specific translation memory (TM).} They report significant overall productivity gains, but with marked differences between the pairs: 59.74\% for the former and only 9.26\% for the latter. Another interesting comparison of HT, PBSMT, and NMT post-editing settings performed on a literary text (chapter from a novel) reports an increase in productivity by 36\% for the NMT based setting over HT \citep{toral2018post}.

We have seen in previous studies that throughputs vary considerably depending on the language pair under the lens \citep{green2013efficacy, laubli2019post}. We now discuss some earlier efforts that have included an Indian language in their experiment.

\citet{shah2015post} conducted an experiment where students post-edited parts of a specialized English language textbook on \textit{bioelectromagnetism} into 7 languages, 3 of them being Indian languages including Hindi. They reported an increase in post-edit time by a factor of 3--5 when the target language was an Indian language. They put this down to greater terminological distance between English and Indian languages compared to other languages in their experiment. They do not study and compare against the HT condition, or report on technical or cognitive indicators.

\citet{carl2016english} also report results on Hindi (amongst 6 languages)  comparing the HT and PE conditions. Their English-Hindi results are based on an existing multilingual translation database that contains experimental data around translators' activities in both conditions. They find in favour of the PE condition across all languages when measured on temporal indicators, but report translating into Hindi to be the slowest amongst the 6. They do not quantify average throughput gain or time savings.

\citet{meetei2020english} compare PE behaviour when translating from English into 3 Indian languages (Manipuri, Mizo and Hindi). They conduct light post-editing and report Hindi to be the fastest to post-edit amongst the three languages.\footnote{It is also often referred to as \textit{good enough} translations and is lower than publishable quality translations.} They ascribe it to the availability of relatively mature MT systems in the English-Hindi direction compared to Mizo and Manipuri, which are low-resource languages. They use student volunteers and do not investigate cognitive indicators.

\citet{ahmad2018transzaar} present an industry perspective and claim a 2--3 fold increase in productivity for users using their tools in combination with MT. However, they base this on longitudinal tracking of their users.

While all these past studies have certainly helped in providing insights into post-editing behaviour in Indian languages in general and in Hindi in particular, we sense a need for a more in-depth look along all PE effort indicators within at least one Indian-language setting. With our current work, we seek to address this gap.

We see our contribution differentiated from previous related work as follows: \textit{(i)} we present in-depth results from all three primary indicators of PE effort (temporal, technical and cognitive) for the English-Hindi direction; \textit{(ii)} we account for per translator and per item variation with the use of mixed-effects models; \textit{(iii)} we utilize professional translators in order to accurately gauge the impact of contrasting conditions; \textit{(iv)} we conduct a human quality-rating exercise comparing target text produced in both conditions; \textit{(v)} we present correlations of automatic MT metrics with PE effort indicators.

\section{Experimental Setup}
\label{setup}

\begin{figure}
  \centering
  \includegraphics[width=0.5\textwidth]{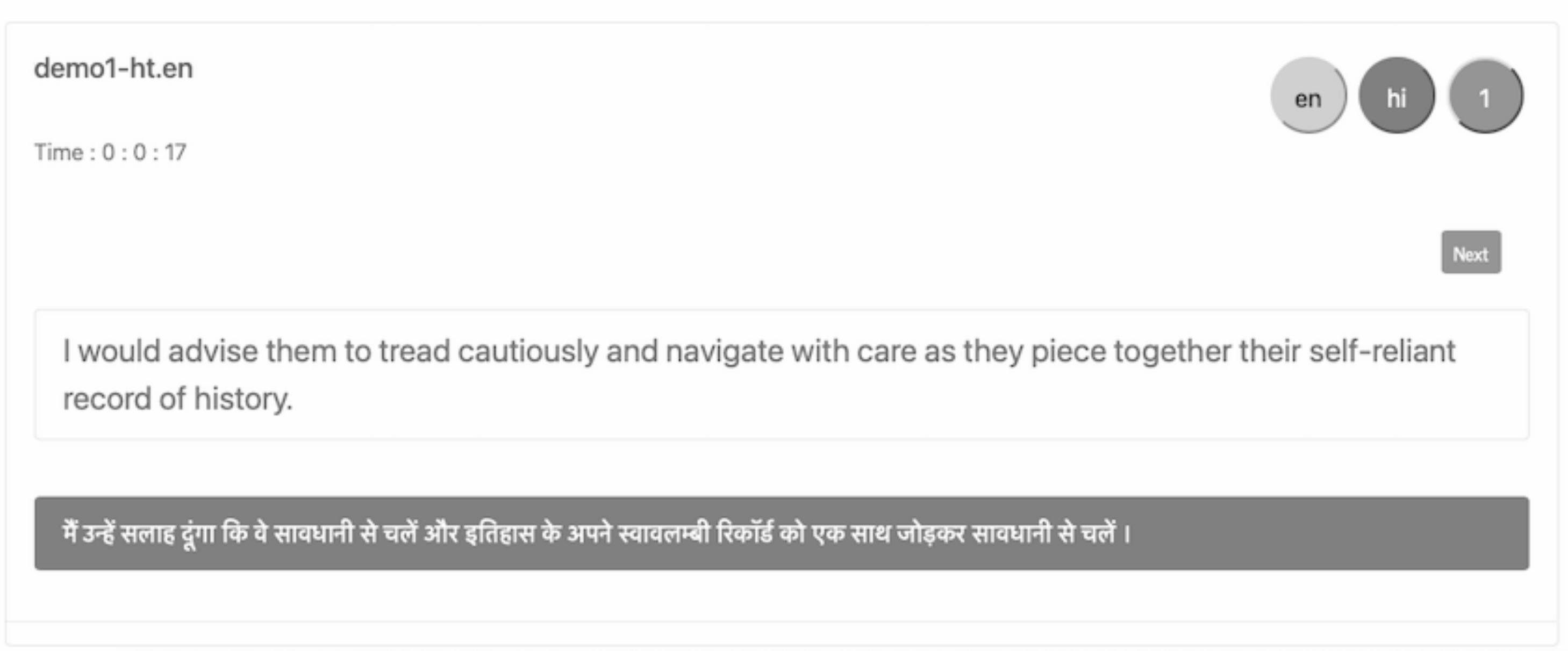}
  \caption{Shows a snapshot of the translator workbench used in the study. Depicts an example in the PE condition pre-filled with an MT proposal; for the HT condition the text area is left blank.}
  \label{fig:interface}
\end{figure}

We conducted this experiment under a 2 (translation conditions) $\times$ 200 (source segments) mixed design. All subjects saw both factor levels (HT and PE), but only one combination for each level as having been exposed to a source segment in one condition would have affected their translation in the other. The study was conducted online over 5 consecutive days with 2 sessions per day. The sessions were not time bound. Subjects translated 2 files of 10 segments each in alternating conditions in each session.

All subjects participated in a warm-up translation exercise a week prior to the start of the actual task. This was done to establish familiarity with the interface used in the study. We chose to adapt an existing beta version of a web-based translation workbench by adding extensive keystroke logging features along with some other minor tweaks.\footnote{https://indictranslate.in/} The UI itself was kept clean and uncluttered, serving one segment at a time to the translators. This meant that while translators had previous context of the text under translation, they could not navigate ahead for context. A timer was displayed once a translator navigated to each new segment. This was meant to prompt the translator to focus on the activity at hand. Figure \ref{fig:interface} shows the workbench interface as seen by translators under the PE condition.

We instructed the participants to aim for publishable translation quality. They were free to conduct web searches and consult online or offline dictionaries, but were discouraged from spending too much time doing so.\footnote{This was done as technical terminology related difficulties have previously been noted for this language direction \citep{shah2015post}.} It was deemed acceptable to transliterate any technical terms or terminology into Hindi if they could not find its translation even with the aid of resources available to them. However, they were strictly prohibited from consulting any online MT engines during the task. Subjects were encouraged to complete each task (consisting of 10 segments) in a single sitting without a break.

Previous studies, such as those discussed earlier, have noted the impact of a number of different variables (language pairs, MT paradigms, text domains, translation environments, translator competencies) on translation throughputs. This calls for not only careful experiment design, but also utilization of techniques that can help with the testing and inference of results. \citet{green2013efficacy} utilized one of the first such designs for post-editing productivity studies and deployed mixed-effects models \citep{baayen2008mixed} to account for inter-language, inter-subject, and inter-item variability.

Mixed-effects models are able to model this variability in two ways: \textit{(i)} through random intercepts, that can account for the differences between translators seen in their differing throughputs (or differences between linguistic items due to the features inherent to them); \textit{(ii)} and through a random slope that accounts for how different subjects may experience the change of condition differently. Accounting for these variabilities allows us to isolate the effect of condition, generalize our findings beyond our sample, and avoid the "language-as-fixed-effect fallacy" \citep{clark1973language}.

In fitting our mixed-effects models we follow a methodology similar to the one described by \citet{baayen2008mixed} and followed by \citet{green2013efficacy} and later \citet{toral2018post}. Maximal models were fit when possible \citep{barr2013random}; in case of convergence failure, a less complex model was fit by successively removing the random slopes of the by-subject and by-segment random effects component. Models thus obtained were compared via likelihood ratio tests. We also refit our final models after filtering data points with residuals deviating more than 2.5 standard deviations. This helps check for the influence of any atypical outliers \citep{baayen2008mixed}. We verify the residual plots for normality and homoscedasticity. We utilized the \texttt{lme4} package in R \citep{lme4new} for all mixed-effects models related analyses.

\subsection{Data}
We assembled a corpus of recent English language news articles from two distinct online sources. The choice of news as a domain was motivated by observations of terminology-related difficulties in more specialized domains, as reported by earlier studies \citep{shah2015post}. Each news article was segmented into sentences using the NLTK library and divided into \textit{blocks} of 10 segments.\footnote{https://www.nltk.org/api/nltk.tokenize.html} Only those blocks were used that fell within a \textit{MEAN $\pm$ SD} of the corpus mean (Table \ref{tab:data}). We prioritized the continuity of a news article across blocks when making \textit{block} selections.\footnote{In 2 cases out of 20 we had to skip the subsequent block, owing to short average sentence lengths of the blocks.} This methodology yielded a total of 200 unique source segments divided into 20 blocks of 10 segments each, spanning 5 different news articles: A1--A5. Conditions were counterbalanced to handle order effects.

\begin{table}
  \scriptsize
\centering
\begin{tabular}{lcccc}
\hline
\textbf{Day} & \textbf{S1-T1} & \textbf{S1-T2} & \textbf{S2-T1} & \textbf{S2-T2}\\
\hline
Day 1 & 20.20 (A1) & 26.30 (A1) & 26.80 (A1) & 21.30 (A2) \\
Day 2 & 20.90 (A2) & 22.30 (A2) & 24.90 (A2) & 23.20 (A3) \\
Day 3 & 26.40 (A4) & 24.60 (A4) & 26.30 (A4) & 22.40 (A4) \\
Day 4 & 22.40 (A4) & 20.30 (A4) & 18.00 (A5) & 18.70 (A5) \\
Day 5 & 22.50 (A5) & 16.00 (A5) & 20.30 (A5) & 23.00 (A5) \\
\hline
\end{tabular}
\caption{Average sentence lengths (in words) per session-task block as presented to translators. Also shown in parenthesis are the source articles used for each block.}
\label{tab:data}
\end{table}

\subsection{Participants}
The participants of our study are self-declared professional translators. We contacted a professional translation service provider to help assemble the  pool.\footnote{http://www.ebhashasetu.com/} A short questionnaire accompanied the registration form for the task. Of our participant pool of 10 subjects, 70\% reported 2--5 years of experience translating in the English-Hindi direction, while 30\% reported 0--2 years of experience. The same percentage breakdown was observed for a question related to previous post-editing experience. All subjects were paid the going market rates for the task regardless of the condition (PE, HT).

\subsection{MT System}
The English$\rightarrow$Hindi MT engine used for the task is a transformer based neural machine translation system. This subword-based NMT system is trained on cleaned WAT 2021 \footnote{http://lotus.kuee.kyoto-u.ac.jp/WAT/indic-multilingual/} English-Hindi training corpus using the Opennmt-py toolkit \cite{klein2020opennmt}. The system also utilizes forward and backward translations on the IndicCorp monolingual corpus to obtain synthetic data for training.\footnote{https://indicnlp.ai4bharat.org/corpora/}  It uses subwords as the basic translation unit with 20,000 merge operations on both source and target languages. The system obtained a BLEU score of \textit{35.46} on cleaned WAT 2021 English-Hindi test data.

\section{Results and Discussion}
\label{results}

\subsection{Pre-processing}
\label{pre-process}
Once we processed the activity logs for all 10 subjects across all 200 segments, they yielded 2000 unique observations. We found that 7 of these items (all from the HT condition) did not contain a final translation, so we discarded those. We think that in these cases the subjects may have accidentally navigated to the next segment without having completed a translation. In the PE condition we found that one subject \textit{P01} had not touched 68\% of the MT outputs and had accepted them without modifications. This was almost 3 times the next highest proportion we detected across all other subjects. We decided to remove all data points generated by this subject. We were thus left with 1793 observations on which we base these results.

We calculated time per segment, source segment lengths (in words and characters), number of keystrokes (total, as well as those belonging to different categories: content, navigation and deletion), average pause duration, initial pause duration, and number of pauses. We also computed H-BLEU \citep{papineni2002bleu}, H-TER \citep{snover2006study} and H-chrF \citep{popovic2015chrf} metrics on the post-edited segments.\footnote{H signifies that scores were computed using the reference generated in the PE condition by the same subject.}

Having assembled this data, we set out to answer the research questions posed earlier in Section \ref{related}.

\subsection{Temporal Effort}
\label{temporal}
We first present a view of temporal effort in terms of productivity measured as words per hour. We see productivity improvements in the PE condition across the board except for subject \textit{P07}. Overall, this translates into a throughput increase from 359 words/hour to 979 words/hour. We thus observe an overall productivity gain of 172\%, which amounts to 63\% in time savings.

This is more than twice the 74\% gain reported by \citep{plitt2010productivity} when studying European-language pairs and the 59.74\% reported by \citep{laubli2019post} recently. But we note that in the first case, the experiments were conducted on PBSMT outputs, and in the second, while NMT was used, the control condition was aided by a TM, thus pushing up the baseline throughputs. With this context in mind, the average productivity gain seen in our study does not appear to be unrealistic.

\begin{figure}
  \centering
  \includegraphics[width=\columnwidth]{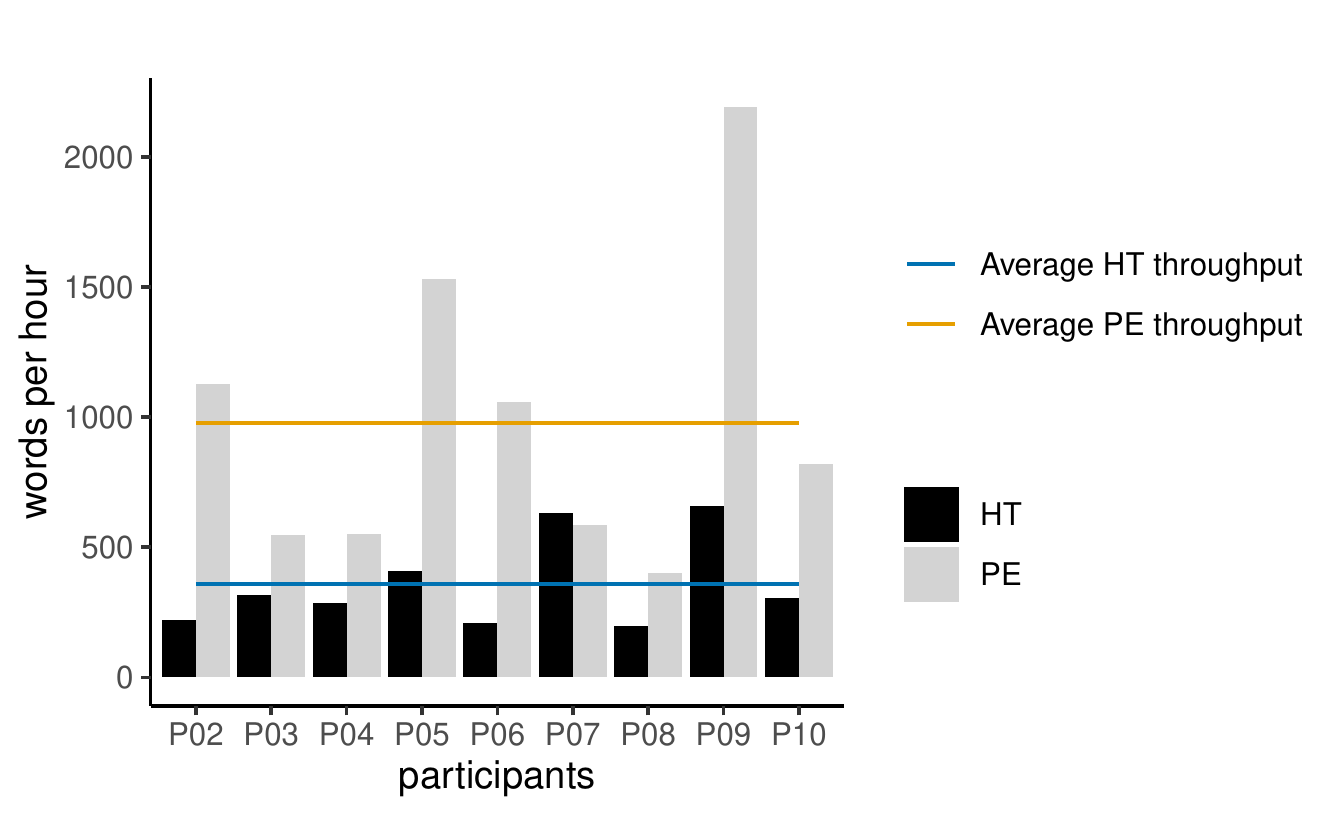}
  \caption{Individual translation throughputs in words per hour and average throughput in each contrasting condition.}
  \label{fig:temporal}
\end{figure}

Figure \ref{fig:temporal} shows a comparison of individual throughputs in contrasting task conditions along with means aggregated for the two conditions.
We also note a great variation in productivity gain amongst subjects ranging from -7\% to 410\%.

Table \ref{tab:gain} helps interpret this further. We contrast the number of unedited and edited MT proposals per subject and their individual productivity gain percentages. It follows that higher the acceptance of MT proposals without modifications by a subject, greater the gain in individual productivity. While this may point to high quality MT output, it also demands a closer scrutiny of the quality of translations generated in each condition. We address this in Section \ref{quality}.

\begin{table}
  \scriptsize
\centering
\begin{tabular}{lcccc}
\hline
\textbf{Participant} & \textbf{Unedited (\%)} & \textbf{Edited (\%)} & \textbf{Productivity Gain (\%)}\\
\hline
P02 & 20 & 80 & 411 \\
P03 & 3 & 97 & 73 \\
P04 & 3 & 97 & 92 \\
P05 & 18 & 82 & 275 \\
P06 & 25 & 75 & 404 \\
P07 & 0 & 100 & -7 \\
P08 & 2 & 98 & 105 \\
P09 & 25 & 75 & 234 \\
P10 & 16 & 84 & 169 \\
\hline
\end{tabular}
\caption{MT segments accepted without modifications and with modifications per subject along with individual productivity gain percentages.}
\label{tab:gain}
\end{table}

We now report the mixed-effects regression results. Plotting temporal data showed a right-skewed distribution, so we log transform all time data before proceeding further. As our goal is to predict translation time and establish the significance of conditions, we fit a linear mixed-effects regression model with two fixed-effect predictors (condition and segment length) and two random-effect predictors (subjects and segments), where on the subject predictor we also include a random-slope for task condition.

In the final model, we observe a significant main effect for both segment length as well as translation condition.\footnote{We utilize the \textit{lmerTest} package that extends results with \textit{p}-values for models built with \textit{lme4}.} Temporal effort significantly increases with segment length, but decreases for the PE condition. Table \ref{tab:significance} shows the significance levels and direction for each predictor in our final models across all PE effort dimensions that we study.

\begin{table*}
  \small
  \centering
  \begin{threeparttable}
\begin{tabular}{l@{\qquad}ccc@{\qquad}ccc}
  \toprule
  \multirow{2}{*}{\raisebox{-\heavyrulewidth}{Predictor}} &  Temporal  & Technical & \multicolumn{3}{c}{Cognitive} \\
  \cmidrule{4-6}
  & & & number & average duration & initial duration \\
  \midrule
  $Segment~length$ & $\uparrow^{***}$ & $\uparrow^{***}$ & $\uparrow^{***}$ & $\uparrow^{***}$ & $\uparrow^{***}$ \\
  $Condition~(PE~vs.~HT)$ & $\downarrow^{***}$ & $\downarrow^{***}$ & $\downarrow^{***}$ & $-$ &  $-$ \\
  \bottomrule
\end{tabular}
\begin{tablenotes}
      \tiny
      \item Significance levels: —($p > 0.1)$, ($p < 0.1)$, \textsuperscript{*}($p < 0.05$),
      \textsuperscript{**}($p < 0.01$), \textsuperscript{***}($p < 0.001$).
      \item Direction: ($\uparrow$ $\downarrow$) arrows depict whether the predictor has a negative or positive correlation with the dependent variable.
    \end{tablenotes}
    \caption{Significance levels of predictors in our final models across all modeled PE effort dimensions.}
\label{tab:significance}
\end{threeparttable}
\end{table*}

\subsection{Technical Effort}
\begin{figure}
  \centering
  \includegraphics[width=\columnwidth]{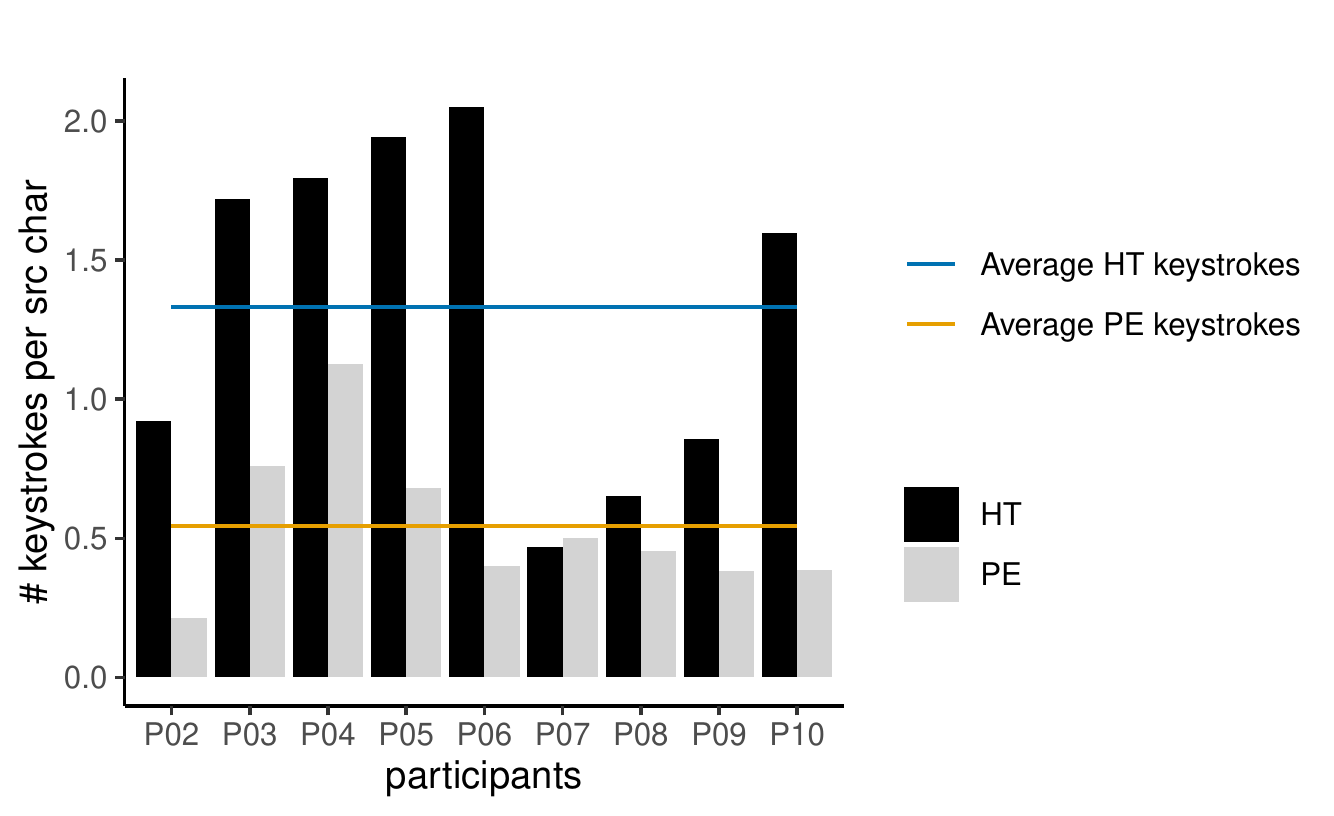}
  \caption{Technical effort estimated as number of keystrokes needed to generate target text per source character.}
  \label{fig:technical}
\end{figure}

We measure technical effort as the number of keystrokes used to generate the target text. We normalize it per source segment character. Figure \ref{fig:technical} shows 1.33 keystrokes used per source character in the HT condition and 0.54 keystrokes in the PE condition, amounting to an effort reduction of 59\%. Contrast this with the 23\% reduction reported by \citet{toral2018post} when post-editing a literary text. Again, except for subject \textit{P07} all participants show reduced effort in the PE condition.

We also classified each keystroke based on the type of the keystroke logged. We classify these into content, navigation, and deletion categories and report the percentage breakdown of the total into these categories in Table \ref{tab:technical-type}. We observe higher navigation and deletion operations in the PE condition (28\% and 26\%) than the HT condition (8\% and 14\%), while content keystrokes register a higher percentage in HT (77\%) compared to PE (46\%). Subject \textit{P08} is an interesting case as they register the highest number of delete operations (they have high navigation numbers too) in either condition amongst all participants. This could point to frequent revisions made on the text.

As number of keystrokes is expressed as counts, we fit a Poisson generalized linear mixed-effects model to predict technical effort. We follow the same methodology as described in the previous section. We again find a significant main effect both for segment length as well as translation condition (Table \ref{tab:significance}), similar to what we saw for the temporal dimension earlier. Technical effort increases with increase in segment length, and decreases for the change in condition to PE.

\begin{table*}
  \scriptsize
  \centering
\begin{tabular}{l@{\qquad}ccc@{\qquad}ccc}
  \toprule
  \multirow{2}{*}{\raisebox{-\heavyrulewidth}{Participant}} & \multicolumn{3}{c}{HT (\%)} & \multicolumn{3}{c}{PE (\%)} \\
  \cmidrule{2-7}
  & Content & Navigation & Deletion & Content & Navigation & Deletion \\
  \midrule
  $P02$ & 81 & 9 & 11 & 41 & 32 & 27 \\
  $P03$ & 94 & 1 & 5 & 58 & 8 & 34 \\
  $P04$ & 86 & 9 & 5 & 48 & 41 & 11 \\
  $P05$ & 85 & 5 & 10 & 56 & 30 & 14 \\
  $P06$ & 90 & 1 & 9 & 54 & 28 & 18 \\
  $P07$ & 66 & 11 & 23 & 56 & 12 & 32 \\
  $P08$ & 24 & 25 & 51 & 10 & 27 & 63 \\
  $P09$ & 78 & 12 & 10 & 35 & 52 & 13 \\
  $P10$ & 94 & 1 & 5 & 59 & 19 & 22 \\
  $Mean \textit{$\pm$ SD}$ & 77.55 $\pm$ 21.80 & 8.15 $\pm$ 7.89 & 14.30 $\pm$ 14.63 & 46.52 $\pm$ 15.91 & 27.69 $\pm$ 13.65 & 25.8 $\pm$ 16.1 \\
  \bottomrule
\end{tabular}
\caption{Types of keystrokes generated by subjects in each condition. }
\label{tab:technical-type}
\end{table*}

\subsection{Cognitive Effort}
\label{cognitive}
Post-editing effort estimation studies based on eye-tracking data use  \textit{fixations} as a proxy to estimate cognitive load; the idea being that greater the number and duration of fixations, greater the cognitive load \citep{o2011towards}. In the absence of eye-tracking data, the use of \textit{pauses} as a proxy for cognitive load is also well established \citep{o2006pauses}. We report on three such cognitive indicators: number of pauses, pause duration, and initial pause duration. Findings related to the first two have been reported in previous post-editing literature \citep{green2013efficacy, toral2018post}.\footnote{There is also an indicator reported as pause ratio which we eschew in favour of initial pause time.} The third (initial pause duration), we introduce in order to gauge differences in reaction times from when a subject first navigates to a new segment displayed in either condition to their first action on it.

We calculate the time difference between two subsequent key events and consider all observations above \textit{1000ms} to be pauses following \cite{o2006pauses, koehn2009process}.

\begin{figure}
  \centering
  \includegraphics[width=\columnwidth]{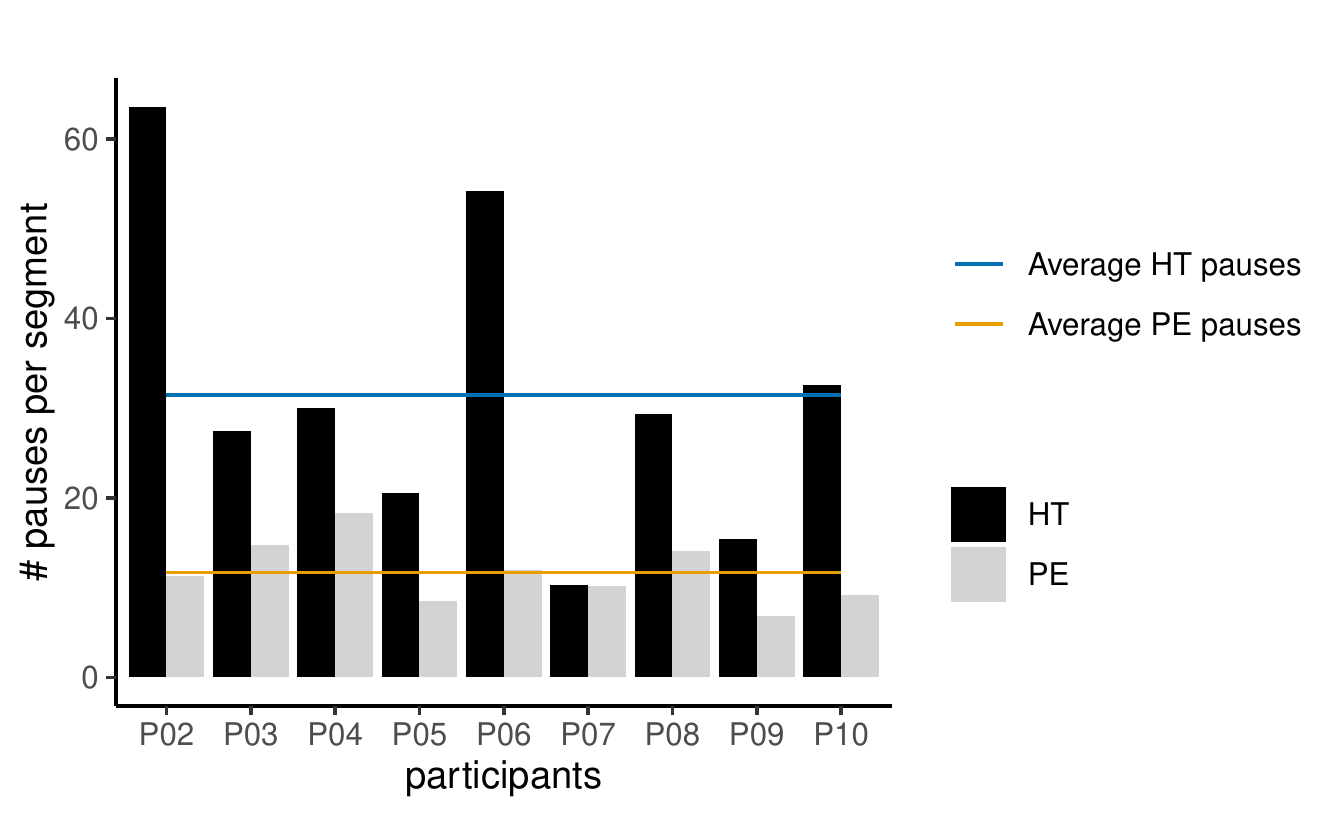}
  \caption{Cognitive effort estimated as average number of pauses per source segment.}
  \label{fig:cognitive-numpause}
\end{figure}

Figure \ref{fig:cognitive-numpause} shows the differences in the frequency of pauses for each subject in the two conditions. We notice a reduction of 63\% in the PE condition from 31 pauses per segment in HT to 12 pauses per segment in PE. This points to a much reduced cognitive load when post-editing.

\begin{figure}
  \centering
  \includegraphics[width=\columnwidth]{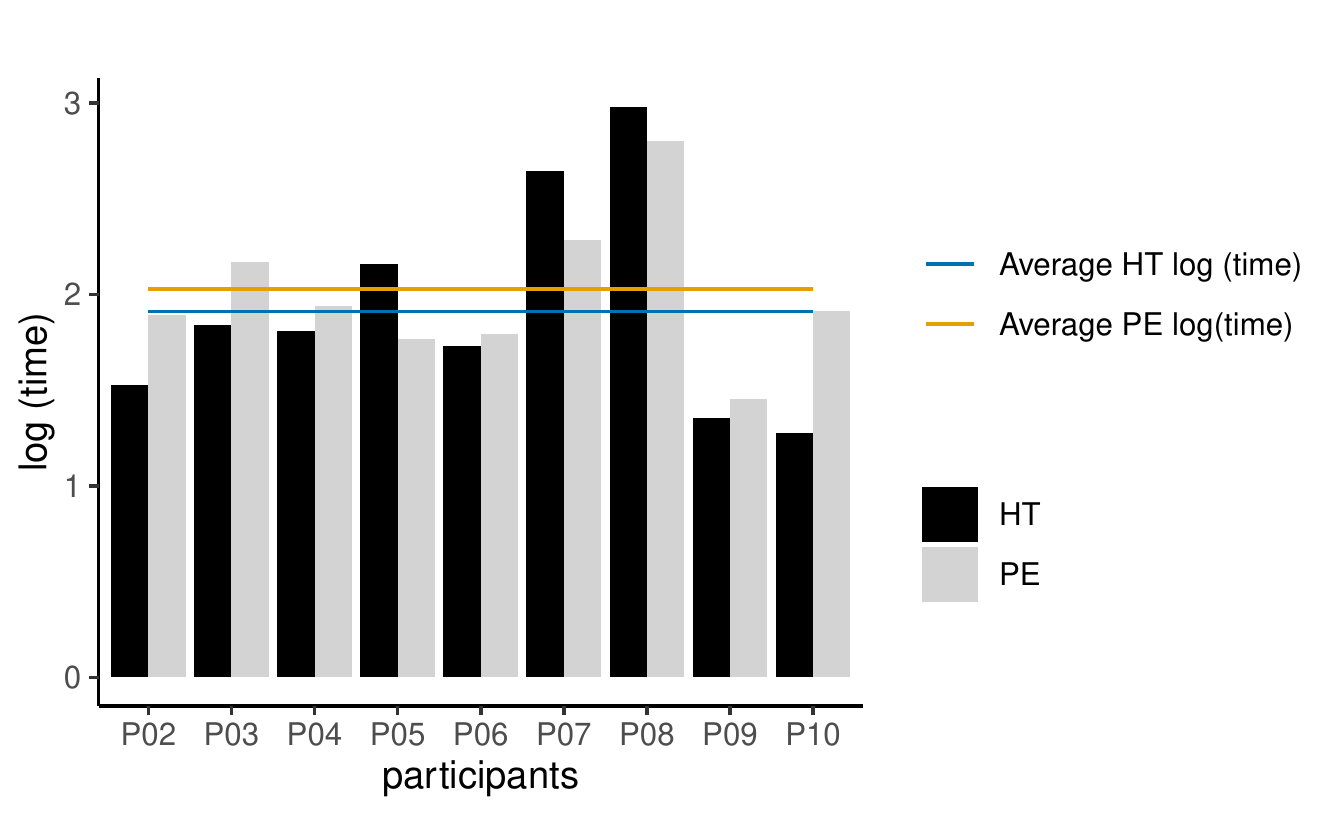}
  \caption{Cognitive effort estimated as average pause duration per source segment.}
  \label{fig:cognitive-pausetime}
\end{figure}

However, a similar exercise on pause duration data reveals an increase of approximately 12\% in the PE condition compared to the HT condition (Figure \ref{fig:cognitive-pausetime}). Although, it is not significant, this is in line with findings reported previously comparing these two specific cognitive indicators \citep{green2013efficacy}.

\begin{figure}
  \centering
  \includegraphics[width=\columnwidth]{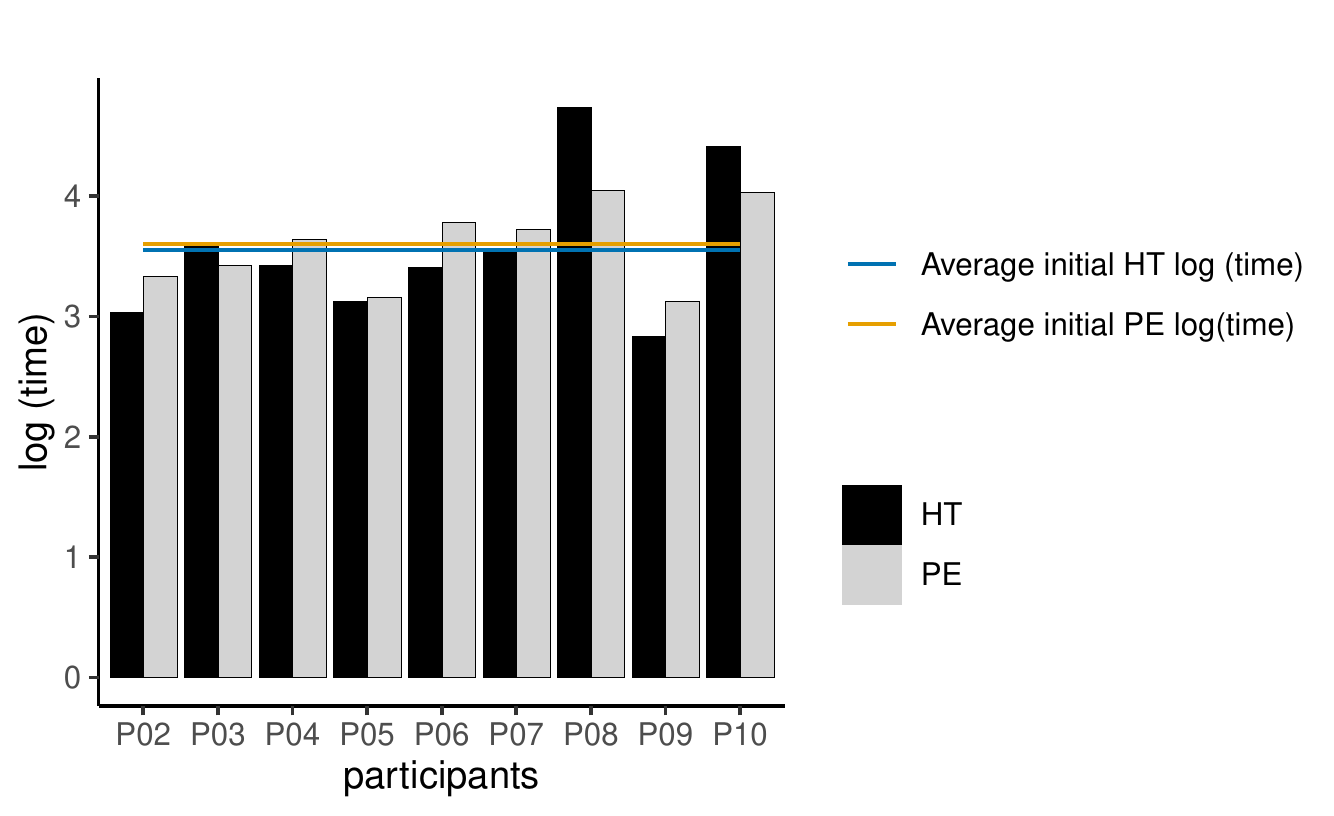}
  \caption{Cognitive effort estimated as average initial pause duration per source segment.}
  \label{fig:cognitive-initial}
\end{figure}

We finally compare initial pause duration between PE and HT. We expect this initial load to be higher for the PE condition given that there are two segments displayed to the subject in this condition: the source segment and the MT proposal, which have to be read and comprehended before starting the post-editing activity. This seems to hold, but not significantly, as we see only a small increase of about 5\% for the PE condition (Figure \ref{fig:cognitive-initial}). One explanation could be that in the PE condition, the MT proposal in spite of registering a higher cognitive load initially also later acts as a helpful prompt for the subject. An eye-tracking based experiment might prove useful in teasing apart these two opposite effects.

When comparing the means\footnote{After transforming back from log scale. Also, note that pause duration does not include initial pause duration as a component, and is the duration of pauses after post-editing starts.} for pause duration and initial pause duration we find pause duration (6.77s for HT and 7.71s for PE) to be considerably lower than initial pause duration (35.02s for HT and 36.67s for PE). The translator therefore, takes a longer initial pause to start formulating a response, but once they start the activity, they take considerably shorter pauses.

We go on to fit three mixed-effects models to validate these findings. A Poisson generalized mixed-effects model to estimate pause counts finds significant main effects for segment length and condition. Cognitive effort (measured as count of pauses) increases with segment length and decreases for the change in condition to PE.

The other two linear mixed-effects models fit to predict average pause duration, and initial pause duration find a significant effect only for segment length and not for condition (Table \ref{tab:significance}). This shows that while cognitive effort certainly increases with segment length, the change in condition to PE, does not have a discernible effect on cognitive effort, when measured by the average and initial time duration indicators.

\subsection{Quality Judgements}
\label{quality}
To evaluate whether the quality of texts created in the PE condition matched those created in the HT condition, we conducted a human judgement based pairwise ranking task \cite{callison2011findings} on a small sample. We randomly sampled 3 target segments per condition from each subject. For each target text thus obtained, we paired it with another random sample after constraining on condition. We thus obtained 60 HT-PE pairs for evaluation. As we discovered issues (discussed earlier in Section \ref{pre-process}) with subject \textit{P01}'s data after the quality evaluation exercise had already been completed, we removed any pairs that had a segment translated by the subject. We report our results on this filtered set that consists of 47 pairs. Five  evaluators were asked to judge each pair. Ties were allowed.

Table \ref{tab:quality} shows the judgements from evaluators represented as win-loss statistics on the PE condition. We notice a high number of ties and a slight preference for the PE condition. However, the preference does not test to be significant on a sign test ignoring ties (\textit{p}-value = 0.08). We conclude that translating in either condition produces similar quality target segments. However, we realise that the sample size was quite small compared to the number of possible combinations across all participants. We hope to conduct a more thorough review of quality in future.

\begin{table}
  \centering
  \scriptsize
\begin{tabular}{c@{\qquad}ccc@{\qquad}ccc}
  \toprule
  \multirow{2}{*}{\raisebox{-\heavyrulewidth}{Evaluator}} & \multicolumn{3}{c}{PE vs. HT}\\
  \cmidrule{2-4}
  & Win & Loss & Tie\\
  \midrule
  $E1$ & 18 & 14 & 15 \\
  $E2$ & 27 & 15 & 5\\
  $E3$ & 7 & 7 & 33\\
  $E4$ & 14 & 11 & 22\\
  $E5$ & 13 & 11 & 23\\
  \bottomrule
\end{tabular}
\caption{Pairwise quality judgements on sampled target texts reported as win, loss, and ties for PE against HT.}
\label{tab:quality}
\end{table}

\subsection{Automatic Quality Metrics}

Finally, we investigate the correlations of some popular automatic MT evaluation metrics with the post-editing effort indicators reported so far in this study.  We generated metric scores by comparing the MT proposal against its post-edited reference. We calculate scores for H-(BLEU, TER, and chrF).

Table \ref{tab:metrics} shows moderate correlations for all 3 MT metrics on the temporal indicator, similar to what \citet{tatsumi2009correlation} also reported for this indicator. Correlations then get stronger on the technical indicator and then fade for the cognitive indicator.

We believe this may be because cognitive effort is the only one out of the three PE dimensions we studied that is not directly observed (instead, inferred from pause frequency and pause duration data), whereas the technical and temporal indicators can be measured more directly. This is similar to findings previously reported by \citet{moorkens2015correlations} who note a similar correlation trend across the three PE effort dimensions. The technical effort indicator appears to be the one most strongly correlated with automatic metrics.

The other two cognitive indicators (average and initial pause duration), which did not test significant as per our mixed-effects models, also do not show any correlation with any of the MT metrics – coefficients obtained were close to 0. We omit reporting them in Table \ref{tab:metrics} due to space constraints.

\begin{table}
  \centering
  \scriptsize
  \begin{threeparttable}
    \begin{tabular}{l@{\qquad}lll@{\qquad}lll}
      \toprule
      \multirow{2}{*}{\raisebox{-\heavyrulewidth}{MT Metric}} & \multicolumn{3}{c}{PE Indicator}\\
      \cmidrule{2-4}
      & temporal & technical & cognitive (\# pauses)\\
      \midrule
  $(H)BLEU$ &   r = $-.56$, $^{***}$ &   r = $-.71$, $^{***}$ &   r = $-.48$, $^{***}$ \\
  $(H)TER$ &   r = $+.54$, $^{***}$ &   r = $+.70$, $^{***}$ &   r = $+.49$, $^{***}$\\
  $(H)chrF$ &   r = $-.56$, $^{***}$ &  r =  $-.73$, $^{***}$ &  r =  $-.49$, $^{***}$\\
  \bottomrule
\end{tabular}
\begin{tablenotes}
\item Note: All coefficients for r(898). For TER lower is better hence the positive correlation. The other two cognitive indicators (average and initial pause duration) did not show any correlation with any of the metrics -- coefficients were close to 0.
\end{tablenotes}
\caption{Correlations of PE Effort indicators with automatic MT metrics.}
\label{tab:metrics}
\end{threeparttable}
\end{table}

\section{Conclusion}
We conducted a post-editing effort assessment study and presented detailed analysis of effort indicators along the temporal, technical and cognitive dimensions. We observed that in the temporal dimension, post-editing reduced translation time by 63\%; in the technical dimension it reduced number of key strokes by 59\%; and in the cognitive dimension, it reduced the frequency of pauses by 63\%. However, it increased average pause duration by 12\% and average initial pause duration by 5\%.

We then compared the quality of translations generated in each condition and found them to be similar.

And finally, we detected moderate to strong  correlations for 3 automatic MT evaluation metrics across all PE effort indicators, with technical effort most strongly correlating with automatic MT metrics.

The last two observations regarding human quality judgement, and MT metrics and their correlations demand a closer look, which was not possible owing to time and space constraints. We expect to undertake this as part of our future work. We also propose to extend this study by including a third condition in future, either as an additional MT engine to check if MT quality differences show up in PE effort indicators \citep{toral2018post}, or by the use of translation aids (TM) to gauge their impact in a similar manner \citep{laubli2013assessing, laubli2019post}.

We also intend to study other language pairs, especially those within the multilingual Indian context.

\label{future}

\bibliography{anthology,icon2021}
\bibliographystyle{acl_natbib}



\end{document}